\documentclass{article}

    \usepackage[preprint]{neurips_2024}

\usepackage[utf8]{inputenc} 
\usepackage[T1]{fontenc}    
\usepackage{hyperref}       
\usepackage{url}            
\usepackage{booktabs}       
\usepackage{amsfonts}       
\usepackage{nicefrac}       
\usepackage{microtype}      
\usepackage{xcolor}         
\usepackage{graphicx}%
\usepackage{multirow}%
\usepackage{amsmath,amssymb,amsfonts}%
\usepackage{amsthm}%
\usepackage{mathrsfs}%
\usepackage[title]{appendix}%
\usepackage{xcolor}%
\usepackage{textcomp}%
\usepackage{manyfoot}%
\usepackage{booktabs}%
\usepackage{algorithm}%
\usepackage{algorithmicx}%
\usepackage{algpseudocode}%
\usepackage{listings}%
\usepackage{amsmath}
\usepackage{amsmath} 
\usepackage{caption} 
\usepackage{subcaption} 
\usepackage{graphicx}
\usepackage{tcolorbox}
\usepackage{caption}
\usepackage{array}
\usepackage{float}
\usepackage{adjustbox}
\usepackage{multirow}
\usepackage{array,booktabs,longtable,tabularx}
\usepackage{float}     
\usepackage{booktabs}     
\usepackage{caption}   

\usepackage{bbm}
\usepackage{graphicx}   
\usepackage{booktabs}   

\usepackage[acronym]{glossaries}
\makeglossaries
\newacronym{nlp}{NLP}{Natural Language Processing}
\newacronym{rag}{RAG}{Retrieval Augmented Generation}
\newacronym{llm}{LLM}{Large Language Models}
\newacronym{epu}{EPU}{Economic Policy Uncertainty}
\newacronym{ml}{ML}{Machine Learning}
\newacronym{chatgpt}{ChatGPT}{Chat Generative Pre-trained Transformer}
\newacronym{ai}{AI}{Artificial Intelligence}
\newacronym{ura}{URA}{Uganda Revenue Authority}
\newacronym{sota}{SoTA}{State of The Art}
\newacronym{gpt}{GPT}{Generative Pre-trained Transformer}
\newacronym{rlhf}{RLHF}{Reinforcement Learning from Human Feedback}
\newacronym{bow}{BoW}{Bag of words}
\newacronym{tfidf}{TF-IDF}{Term Frequency Inverse Document Frequency}
\newacronym{word2vec}{Word2Vec}{Word to Vector}
\newacronym{doc2vec}{Doc2Vec}{Document to Vector}
\newacronym{bert}{BERT}{Bidirectional Encoder Representations from Transformers}
\newacronym{roberta}{RoBERTa}{Robustly Optimized BERT Pretraining Approach}
\newacronym{sbert}{SBERT}{Sentence Bidirectional Encoder Representations from Transformers}
\newacronym{svm}{SVM}{Support Vector Machines}
\newacronym{rnn}{RNN}{Recurrent Neural Network}
\newacronym{gru}{GRU}{Gated Recurrent Unit}
\newacronym{lstm}{LSTM}{ Long Short Term Memory}
\newacronym{electra}{ELECTRA}{Efficiently Learning an Encoder that Classifies Token Replacements Accurately}
\newacronym{xlnet}{XLNeT}{Extra Long Network}
\newacronym{bloom}{BLOOM}{BigScience Large Open-science Open-access Multilingual Language Model}
\newacronym{t5}{T5}{Text-to-Text Transfer Transformer}
\newacronym{opt}{OPT}{ Open Pretrained Transformer}
\newacronym{bart}{BART}{ Bidirectional and Auto-Regressive Transformer}
\newacronym{llama}{LLAMA}{Large Language Model Meta AI}
\newacronym{cot}{CoT}{Chain of Thought}
\newacronym{tot}{ToT}{Tree of Thought}
\newacronym{react}{ReACT}{Reasoning + Acting}
\newacronym{instructgpt}{InstructGPT}{Instruction Tuned Generative Pretrained Transformer}
\newacronym{lm}{LM}{Language Model}
\newacronym{tf}{TF}{Term Frequency}
\newacronym{idf}{IDF}{Inverse Document Frequency}
\newacronym{cbow}{CBOW}{Continuous Bag of Words}
\newacronym{glove}{GloVe}{Global Vectors for Word Representation}
\newacronym{rbf}{RBF}{Radial Basis Function}
\newacronym{gbm}{GBM}{Gradient Boosting Machines}
\newacronym{relu}{ReLU}{Rectified Linear Unit}
\newacronym{ce}{CE}{Cross Entropy}
\newacronym{pe}{PE}{Positional embeddings}
\newacronym{mlm}{MLM}{Masked Language Modeling}
\newacronym{ntp}{NTP}{Next Token Prediction}
\newacronym{nsp}{NSP}{Next Sentence Prediction}
\newacronym{roots}{ROOTS}{Responsible Open-science Open-access Text Sources}
\newacronym{c4}{C4}{Colossal Clean Crawled Corpus}
\newacronym{flan}{FLAN}{ Fine-tuned LAnguage Net}
\newacronym{palm}{PaLM}{ Pathways Language Model}
\newacronym{qa}{QA}{Question Answering}
\newacronym{dpo}{DPO}{Direct Preference Optimization }
\newacronym{bm25}{BM25}{Best Matching 25}
\newacronym{api}{API}{Application Programming Interface}
\newacronym{clip}{CLIP}{Contrastive Language Image Pretraining}
\newacronym{lda}{LDA}{Latent Dirichlet Allocation}
\newacronym{cnn}{CNN}{Convolution Neural Network}
\newacronym{knn}{KNN}{K-Nearest Neighbors}
\newacronym{automl}{AutoML}{Automated Machine Learning}
\newacronym{bilstm}{ BiLSTM}{Bidirectional Long Short Term Memory}
\newacronym{it}{IT}{Information Technology}
\newacronym{albert}{ALBERT}{A Lite BERT}
\newacronym{distilbert}{DistilBERT}{Distilled  BERT}
\newacronym{stm}{STM}{Structural Topic Model}
\newacronym{oecd}{OECD}{Organisation for Economic Co-operation and Development}
\newacronym{dp}{DP}{Data Programming}
\newacronym{wawa}{WAWA}{Worker Agreement with Aggregate}
\newacronym{mistral_24b}{Mistral-24B}{Mistral 24 Billion parameters}
\newacronym{qwen2_7b}{Qwen2-7B}{Qwen version two with 7 billion parameters}
\newacronym{llama3_8b}{LLAMA3-8B}{LLAMA version three with 8 billion parameters}
\newacronym{mistral3_7b}{Mistral3-7B}{Mistral version two with 7 billion parameters}
\newacronym{qwen2_14b}{Qwen2-14B}{Qwen version two with 14 billion parameters}
\newacronym{emr}{EMR}{Exact Match Ratio}
\newacronym{lrp}{LRP}{Label Ranking Precision}
\newacronym{gm}{GM}{Geometric Mean}
\newacronym{auc}{AUC}{Area Under the Curve}
\newacronym{auprc}{AUPRC}{Area Under the Precision Recall Curve}
\newacronym{ck}{CK}{Cohen’s Kappa}
\newacronym{mcc}{MCC}{Matthews Correlation Coefficient}
\newacronym{ws}{WS}{weak supervision}
\newacronym{roc}{ROC}{Receiver Operating Curves }
\newacronym{usa}{USA}{United States of America}
\newacronym{gpu}{GPU}{Graphics Processing Unit}
\newacronym{distilroberta}{DistilRoBERTa}{Distiled RoBERTa}
\newacronym{opt_125m}{OPT-125M}{OPT with 125 million parameters}
\newacronym{opt_350m}{OPT-350M}{OPT with 350 million parameters}
\newacronym{qwen2_3b}{Qwen2-3B}{Qwen version two with 3 billion parameters}
\newacronym{qwen2_0.5b}{Qwen2-0.5B}{Qwen version two with 0.5 billion parameters}
\newacronym{llama3_1b}{LLAMA3-1B}{LLAMA version three with 1 billion parameters}
\newacronym{llama3_3b}{LLAMA3-3B}{LLAMA version three with 3 billion parameters}
\newacronym{llama2_7b}{LLAMA2-7B}{LLAMA version two with 7 billion parameters}
\newacronym{mistral2_7b}{Mistral2-7B}{Mistral version two with 7 billion parameters}
\newacronym{qlora}{ QloRA}{ Quantized Low-Rank Adaptation}
\newacronym{rouge}{ROUGE}{Recall Oriented Understudy for Gisting Evaluation}
\newacronym{bertscore}{ BERTScore}{Bidirectional Encoder Representations from Transformers Score}
\newacronym{tin}{TIN}{Tax Identification Number}
\newacronym{vat}{VAT}{Value Added Tax}
\newacronym{zephyr_7b}{Zephyr-7B}{Zephyr model with 7 billion parameters}
\newacronym{hse}{HSE}{Zephyr model with 7 billion parameters}
\newacronym{colbert}{ColBERT}{Contextualized Late Interaction over BERT}
\newacronym{maxsim}{MaxSim}{Maximum Cosine Similarity}
\newacronym{mt}{MT}{Machine Translation}
\newacronym{dm}{DM}{Definition Modeling}
\newacronym{pg}{PG}{Paragraph Generation}
\newacronym{blue}{BLEU}{Bilingual Evaluation Understudy}
\newacronym{llava_7b}{LLAVA-7B}{LLAVA with 7 billion parameters}
\newacronym{llava_13b}{LLAVA-13B}{LLAVA with 13 billion parameters}
\newacronym{llama3_70b}{LLAMA3-70B}{LLAMA version three with 70 billion parameters}
\newacronym{llama2_13b}{LLAMA2-13B}{LLAMA version two with 13 billion parameters}
\newacronym{gpt4}{GPT-4}{GPT version four}
\newacronym{gpt40_mini}{GPT-40-mini}{GPT version four mini}
\newacronym{gpt3.5_turbo}{GPT-3.5-Turbo}{GPT version 3.5 Turbo}
\newacronym{cso}{CSO}{Central Statistics Office}
\newacronym{irish}{IRISH}{Documents from Irish publications website}
\newacronym{mistral2_8_x_7b}{Mistral2-8X7B}{Mistral2-8X7B}
\newacronym{llama2_70b}{LLAMA2-70B}{LLAMA version two with 70 billion parameters}
\newacronym{shroom}{SHROOM}{Shared-task on Hallucinations and Related Observable Overgeneration Mistakes}
\newacronym{majority_vote}{Majority Vote}{Majority Vote}
\newacronym{faiss}{FAISS}{Facebook AI Similarity Search}
\newacronym{ground}{Grd}{Groundednes}
\newacronym{rel}{Rel}{Relevance}
\newacronym{stand}{Stand}{Standalone}

\title{Beyond Keywords: Leveraging Generative LLMs and Label Aggregation to Classify Economic Policy Uncertainty in News Articles}

\author{%
  Paul Trust \\
  \texttt{trustpaul1000@gmail.com} \\
}

\begin{document}

\maketitle

\begin{abstract}
This research describes the adaptation of \acrlong{llm} (\acrshort{llm}s) for economic monitoring in the public sector to automatically determine whether an article discusses \acrlong{epu} (\acrshort{epu}) and to identify its specific type.
Previous studies either rely on keywords, which often result in a high count of false positives, or use \acrlong{ml} (\acrshort{ml}) approaches that require a large number of quality human labeled data that is costly and time consuming to acquire. In this study, we propose approaches based on weak supervision techniques, using generative \acrshort{llm}s to create synthetic labels through prompting, making the approach both cost-effective and scalable. Additionally, we propose methods for for multi-label and hierarchical classification of articles related to \acrshort{epu}.
\end{abstract}

\section{Introduction}\label{sec:introduction}

The growing availability of digital text, from news media to online platforms and institutional records, has created new opportunities for studying economic and political dynamics at scale. Computational analysis of such unstructured text has become an important tool in disciplines including economics, sociology, and political science \cite{gentzkow2019text, grimmer2013text}. In economics, textual indicators have been derived from central bank communications \cite{rybinski2019machine}, corporate earnings calls \cite{keith2019modeling}, news articles \cite{baker2016measuring} and social media data \cite{indaco2020twitter}. A well known application is the \acrshort{epu} index by Baker et al. (2016), which measures the frequency of policy- and uncertainty-related terms in news articles \cite{baker2016measuring}. While influential, this keyword-based method is prone to false positives and fails to capture policy context with precision.

To overcome these limitations, a range of \acrshort{ml} techniques have been explored. Unsupervised approaches such as topic modeling and embedding-based expansions provide improved coverage of relevant content \cite{azqueta2017developing, azqueta2020economic, azqueta2023sources, kaveh2021measuring, nyman2020text, wang2024impacts, miranda2023word, jiang2023can, xu2023normalized}, but they remain less accurate than supervised methods. Traditional supervised \acrshort{ml} and more recent deep learning techniques \cite{lolic2022economic, tobback2018belgian, sadorsky2022using, wang2022measuring, rao2024role, chen2024semantics, qureshi2021using, keith2020uncertainty, chang2024understanding, audrino2024quantifying, yeh2024automation, ito2024content} have improved predictive performance but rely on costly human-labeled datasets. This dependence on manual annotation constrains scalability and limits adaptation to new time periods, countries, or policy domains. 
The most closely related work is by Paul et al. (2023), which uses labeling functions to create synthetic labels for training \acrshort{llm}s. However, these generated labels are often unstable, and designing high-quality labeling functions requires substantial domain expertise \cite{trust2023understanding}. Furthermore, most prior research has focused only on binary classification, identifying whether an article discusses \acrshort{epu} without distinguishing between specific types of policy uncertainty.

This research proposes a weakly supervised framework for \acrshort{epu} classification that addresses these challenges. \acrshort{llm}s are prompted to generate synthetic labels, which are then refined using aggregation methods such as \acrlong{dp} (\acrshort{dp}), Dawid–Skene, \acrlong{wawa} (\acrshort{wawa}), and \acrshort{majority_vote}. The resulting labels are used to fine-tune \acrshort{llm}s: \acrshort{bert}, \acrshort{roberta}, and \acrlong{lstm} (\acrshort{lstm}), thereby reducing dependence on human annotation. Beyond binary classification, we extend the task to multi-label and hierarchical settings, enabling the assignment of articles to specific policy subcategories once identified as \acrshort{epu}-related. 
The remainder of this paper is structured as follows: Section \ref{sect:related_work} reviews related work, Section \ref{sect:methods} presents the methodology, Section \ref{sec: results} reports the experimental findings, and Section \ref{section: conclusion} presents the conclusion of this work.

\section{Related Work}\label{sect:related_work}
This section presents related work on \acrshort{epu} classification, categorized into unsupervised \acrshort{ml} methods, traditional \acrshort{ml} approaches, deep learning techniques, and \acrshort{llm} methods, summarized in Table \ref{tab:epu_studies} and discussed next.

\begin{table}[h]
\centering
\caption{Relevant studies on automated methods for \acrshort{epu} classification}
\label{tab:epu_studies}
\renewcommand{\arraystretch}{1.2} 
\setlength{\tabcolsep}{1em} 
\begin{tabularx}{\textwidth}{|l|X|}
\toprule
\textbf{Category} & \textbf{Relevant Works} \\
\midrule
\textbf{Unsupervised \acrshort{ml}} 
& \cite{azqueta2020economic}, \cite{xu2023normalized}, \cite{azqueta2017developing}, \cite{wang2024impacts}, \cite{miranda2023word}, \cite{kaveh2021measuring}, \cite{nyman2020text}, \cite{azqueta2023sources}, \cite{jiang2023can}, \cite{baker2016measuring} \\
\midrule
\textbf{Traditional \acrshort{ml}} 
& \cite{lolic2022economic}, \cite{tobback2018belgian}, \cite{sadorsky2022using}, \cite{wang2022measuring} \\
\midrule
\textbf{Deep Learning \& \acrshort{llm}s} 
& \cite{rao2024role}, \cite{chen2024semantics}, \cite{qureshi2021using}, \cite{keith2020uncertainty}, \cite{chang2024understanding}, \cite{audrino2024quantifying}, \cite{yeh2024automation}, \cite{ito2024content} \\
\bottomrule
\end{tabularx}
\end{table}

\textbf{Unsupervised ML Approaches}: Most \acrshort{epu} detection methods have utilized unsupervised learning techniques to identify news articles related to \acrshort{epu}. The earliest and most widely used approach, introduced by Baker et al. (2016), relies on keyword matching for classification. Building on this, alternative approaches have been proposed to enhance the detection process. For instance, Kaveh et al. (2021) constructed an \acrshort{epu} index by leveraging a word embedding representation space to refine keyword selection for \acrshort{epu} detection in news articles. Nyman and Ormerod (2020) further expanded the original uncertainty keyword list proposed by Baker et al. (2016) using nearest neighbor embeddings and examined the Granger causality between their expanded keyword set and the existing \acrshort{epu} index \cite{nyman2020text}.

While keyword-based methods offer a simple and interpretable solution, they often suffer from high false positive rates and limited semantic understanding. To overcome these limitations, topic modeling techniques have been proposed as a more robust alternative for uncovering latent themes related to \acrshort{epu}. Relevant studies include: Azqueta et al. (2020), who applied topic models to construct \acrshort{epu} indices for European countries \cite{azqueta2017developing, azqueta2020economic, azqueta2023sources}.

Wang et al. (2024) used \acrlong{lda} (\acrshort{lda}) to extract themes from newspapers articles and constructed the Chinese \acrshort{epu} index from them \cite{wang2024impacts}. Miranda et al. (2023) implemented topic modeling with semantic clustering based on word embeddings to detect \acrshort{epu} in digital news \cite{miranda2023word}. Jiang et al. (2023) applied BERTopic and regularized linear models to predict \acrshort{epu} using social media data from Reddit \cite{jiang2023can}.
Xu et al. (2023) utilized an unsupervised approach combining principal component analysis  and random matrix theory to construct \acrshort{epu} indices \cite{xu2023normalized}. 
Although topic modeling captures latent semantic structures, its unsupervised nature limits classification accuracy compared to supervised learning. To improve precision and reliability in \acrshort{epu} detection, traditional \acrshort{ml} methods using labeled data have been developed.

\textbf{Traditional \acrshort{ml} Approaches}: 
Relevant studies in this category include: Lolić et al. (2022)  who applied \acrshort{ml} techniques: linear regression, random forests, \acrlong{gbm} (\acrshort{gbm}), decision trees, extreme gradient boosting, and their ensembles for \acrshort{epu} classification \cite{lolic2022economic}. Tobback et al. (2018) used \acrlong{svm} (\acrshort{svm}) to classify news articles related to \acrshort{epu} from Belgian news sources \cite{tobback2018belgian}. 
Sadorsky (2022) employed various \acrshort{ml} models such as random forests, extremely randomized trees, \acrshort{gbm}, and Naïve Bayes for \acrshort{epu} classification \cite{sadorsky2022using}. Wang et al. (2022) utilized  neural networks, \acrshort{gbm}, a dictionary-based approach, and ensemble classifiers, to classify \acrshort{epu} in Chinese news articles \cite{wang2022measuring}.
The reliance of traditional \acrshort{ml} on manual feature engineering and limited capacity to capture complex language patterns has motivated a shift towards deep learning models and \acrshort{llm}s for more advanced text understanding.

\textbf{Deep Learning \& \acrshort{llm} Approaches}: Representative studies include: Rao et al. (2024) who used \acrshort{lstm}s, \acrshort{gru}s, and multi-layer perceptrons to classify \acrshort{epu} in news articles \cite{rao2024role}. Chen et al. (2024) applied word embeddings based on Skip-Gram along with \acrshort{gru}, \acrlong{cnn} (\acrshort{cnn}), and capsule networks for \acrshort{epu} classification tasks \cite{chen2024semantics}. 
Qureshi et al. (2021) utilized \acrshort{bert} models, keyword extraction algorithms, and a combination of \acrshort{roberta} and sentence transformers to extract and classify \acrshort{epu} related keywords \cite{qureshi2021using}. Keith et al. (2020) combined keyword matching, supervised \acrshort{ml} techniques (logistic regression with \acrlong{bow} (\acrshort{bow}) and \acrshort{bert} embeddings), and prevalence estimation techniques such as probabilistic classify and count, and implicit likelihood to classify news articles related to \acrshort{epu} \cite{keith2020uncertainty}. 

Chang et al. (2024) leveraged \acrshort{bert} to classify trade policy uncertainty based on texts from  management discussion sections of annual reports from Chinese public firms \cite{chang2024understanding}. Audrino et al. (2024) adapted various \acrshort{llm}s, including \acrshort{gpt}-4, \acrshort{llama}, and \acrshort{zephyr_7b} for \acrshort{epu} classification \cite{audrino2024quantifying}. Yeh et al. (2024) applied \acrshort{llm}s such as \acrshort{gpt}-3.5-Turbo, \acrshort{gpt}-4, and Claude3 to generate keywords for \acrshort{epu} classification \cite{yeh2024automation}. 
Ito et al. (2024) fine-tuned \acrshort{llm}s to detect \acrshort{epu}, contributing to the development of Japan's monetary policy index for the period between 2015 and 2016 \cite{ito2024content}. Paul et al. (2023) proposed to use labeling functions to generate synthentic labels to pre-train \acrshort{llm}s \cite{trust2023understanding}.
Although these methods also used \acrshort{llm}s like the approaches proposed in this research, their reliance on costly human-curated labels limits their applicability.

\textbf{Research Gap}: Numerous previous studies  discussed focused on unsupervised \acrshort{ml} approaches, primarily using topic models and keyword matching. While these methods are simple and do not require training, they are less effective than supervised \acrshort{ml} techniques. In contrast, supervised \acrshort{ml} methods, including traditional \acrshort{ml} and deep learning, rely on human-curated datasets, making them less scalable. Constructing a new index for a different year, country, or region requires training new models and generating new labeled datasets, which is resource-intensive. Furthermore, prior research has been limited to binary classification, identifying only whether an article contains \acrshort{epu} or not, with no existing studies attempting to classify specific types of \acrshort{epu} within an article.

In this research, we present our proposal to use \acrshort{llm}s for classifying news articles containing \acrshort{epu}. Instead of relying on human-curated datasets, which are costly to obtain, it proposes a weakly supervised learning approach. \acrshort{llm}s are leveraged to generate training labels or directly classify news articles through prompting. The labels obtained via prompting can either serve as the final classification labels or be refined using ensemble aggregation techniques with multiple \acrshort{llm}s to reduce label noise. Furthermore, this research proposes extending beyond binary classification by adapting \acrshort{llm}s for multi-label and hierarchical classification of \acrshort{epu}-related news articles, an aspect not previously explored in existing research.

\section{Methodology}\label{sect:methods}
This section describes the methodology used for analyzing and classifying news articles related to \acrshort{epu}. The process begins with \acrshort{epu} classification, where weakly supervised labels generated by \acrshort{llm}s are compared with human annotated labels. Next, multi-label text classification is applied to identify different types of \acrshort{epu} present in the articles, with labels assigned only to articles classified as \acrshort{epu}-related. Since this classification follows a hierarchical structure, additional experiments are conducted to assess whether incorporating this hierarchy enhances classification performance.

\subsection{Weakly supervised EPU classification}
For the weakly supervised approach used for \acrshort{epu} classification, we consider a dataset $D^{U}$ of \(n\) unlabeled data points: \(D^{U} = \{x_{i}\}_{i=1}^{n}\). The goal of text classification is to assign each data point \(x_{i}\) to a label $y_{i}$  where each label \(y_{i}\) belongs to one of \(m\) possible classes: $C = \{c_{1}, \ldots, c_{m}\}$, resulting in a labeled dataset: \(D = \{(x_{i}, y_{i})\}_{i=1}^{n}\). 
Unlike traditional supervised learning, which relies on large volumes of human labeled data, weak supervision relies on  synthetic labels generated by \acrshort{llm}s to train the classifiers.
However, a human labeled dataset is used during evaluation to accurately assess the model’s performance.

We formulate the \acrshort{epu} classification as a binary classification task with two target classes: \(C = \{\textit{``No \acrshort{epu}"}, \textit{``\acrshort{epu}"}\}\). The proposed approach comprises three main stages. In the initial stage, \acrshort{llm}s are used to generate labels via prompting, guided by instructions embedded in the prompt illustrated in Figure~\ref{fig:data_generation_prompt_weak}. This prompt design is inspired by the annotation guidelines developed by Baker et al. (2016) for human annotators \cite{baker2016measuring}.

While the work by Paul et al. (2023) \cite{trust2023understanding}, which this research builds on employed multiple labeling functions such as keyword lookups and semantic similarity, the results presented here are based solely on \acrshort{llm}-based prompting. A single prompt (Figure~\ref{fig:data_generation_prompt_weak}) is used to encode the labeling logic, serving as a soft rule-based alternative that simplifies the process and leverages recent advances in \acrshort{llm} capabilities.

As a result, a label matrix is automatically constructed, where each news article receives multiple weak labels from \(k\) \acrshort{llm}s represented as: \(\lambda = \{\lambda_{1}, \ldots, \lambda_{k}\}\). For a given \acrshort{llm} \(\lambda_{j}\), the output is \(\lambda_{j}(x_{i}) \in \{0,1\}\), where \(1\) corresponds to an \(\textit{``\acrshort{epu}"}\) label and \(0\) corresponds to  a \(\textit{``No \acrshort{epu}"}\) label.
Since different \acrshort{llm}s may assign different labels to the same input \(x_{i}\), the second stage focuses on aggregating the \(k\) generated synthetic labels to produce a less noisy single label. To accomplish this, various ensemble methods are employed which are detailed in Section \ref{section:aggregation}.

\begin{figure}[H]
\begin{tcolorbox}[
    colback=white, 
    colframe=black, 
    sharp corners=south, 
    boxrule=0.5mm, 
    arc=3mm, 
    title=Label Generation Prompt
]
You are a classifier that determines whether a given news article or passage discusses \textbf{Economic Policy Uncertainty (EPU)}.\\[1ex]

Your decision process for each passage should be:\\
1. Check if the article mentions \textbf{general economic uncertainty}.\\
\hspace*{1em} - Examples include: job market fluctuations, inflation concerns, recession fears.\\
\hspace*{1em} - If the article only discusses general economic uncertainty but does not link it to government policy, classify as \textbf{EPU=0}.\\[1ex]

2. Check if the article explicitly connects economic uncertainty to \textbf{government policy actions, decisions, or inactions}.\\
\hspace*{1em} - Mentions of policy-related economic uncertainty in the past, present, or future all qualify.\\
\hspace*{1em} - If this condition is met, classify as \textbf{EPU=1}.\\[1ex]

Further indicators that qualify for \textbf{EPU=1} classification include uncertainty related to policy decisions in any of the following areas:\\
\hspace*{1em} - Monetary policy (e.g., central bank actions, interest rates, money supply)\\
\hspace*{1em} - Fiscal policy (e.g., government spending, tax reforms)\\
\hspace*{1em} - Regulations (e.g., labor laws, environmental policies, financial regulations)\\
\hspace*{1em} - Trade policy (e.g., tariffs, trade agreements, export/import restrictions)\\
\hspace*{1em} - Government leadership uncertainty (e.g., elections, political conflicts)\\
\hspace*{1em} - Sovereign debt and currency stability (e.g., exchange rate policies, foreign reserves)\\
\hspace*{1em} - National security and terrorism (if linked to economic uncertainty)\\
\hspace*{1em} - Social safety nets and entitlement programs (e.g., welfare, healthcare policies)\\[1ex]

\textbf{Output:}\\
- If policy-related economic uncertainty is present → \textbf{EPU=1}.\\
- Otherwise → \textbf{EPU=0}.\\
\end{tcolorbox}
\caption{\acrshort{epu} label generation prompt.}
\label{fig:data_generation_prompt_weak}
\end{figure}

\subsubsection{Weak Label Aggregation}\label{section:aggregation}
In this stage, various aggregation methods including: \acrlong{dp} (\acrshort{dp}), \acrshort{majority_vote}, \acrlong{wawa} (\acrshort{wawa}), and the Dawid-Skene model, are used to consolidate multiple noisy labels generated by different \acrshort{llm}s into a single label.

\begin{itemize}    

\item \textbf{\acrshort{majority_vote}}:
\acrshort{majority_vote} \cite{penrose1946elementary} is a simple yet powerful model often used in ensemble learning techniques. Consider a setting with \( k \) classifiers  in our case the output of $k$ \acrshort{llm}s on the same input text, each providing a vote for one of the possible classes \( c_i \in C \) in a classification task. The \acrshort{majority_vote} is determined by aggregating these votes and selecting the class with the highest number of votes.

If we let \( v_{jc_i} \) to be a binary indicator where \( v_{jc_i} = 1 \) if classifier \( \lambda_{j} \) votes for class \( c_i \), and \( v_{jc_i} = 0 \) otherwise. The total number of votes for class \( c_i \) is then given by: \( V_{c_i} = \sum_{j=1}^{k} v_{jc_i} \). The predicted class \( \hat{y} \) is the one with the maximum \( V_{c_i} \) shown in Equation \ref{eqn:majority_weak}.

\begin{equation}
\label{eqn:majority_weak}
     \hat{y} = \arg\max_{c_i} V_{c_i}
\end{equation}
This method is based on the idea that combining outputs from multiple classifiers can produce more accurate results, especially when each classifier brings different and complementary strengths to the task.

\item \textbf{\acrshort{dp}}: \acrshort{dp} \cite{snorkel2017} is an aggregation method that assigns a \textit{noisy} label to each news article in an unsupervised learning manner. The outputs generated by \acrshort{llm}s through prompting can be viewed as multiple annotators labeling the same article, similar to a crowd-sourcing setup \cite{mishra2021cross}. These labels may contain conflicts and correlations, even when assigned by domain experts.  

Instead of a simple \acrshort{majority_vote}, \acrshort{dp} employs a probabilistic framework to capture the structure and relationships within the label matrix. The probabilistic model \( P_{\theta}(\Lambda,Y) \) is defined as the joint probability of the \acrshort{llm}s' outputs (label matrix) \( \Lambda \) and the latent (unobserved) true class labels \( Y \). The label matrix is encoded using a factor model with three factor types representing conflicts, correlations, and propensity (instances where \acrshort{llm}s generate valid labels).  

The generative model is expressed in Equation  \ref{eqn:snorkel_gen_eqn_1}.
\begin{equation}\label{eqn:snorkel_gen_eqn_1}
  P_{\theta}(\Lambda,Y)=Z_{\theta}^{-1}exp(\sum_{i=1}^{n}\theta^{T}\phi_{i}(\Lambda _{i},y_{i}))
\end{equation}
where \( Z_{\theta} \) is the normalizing constant, \( \phi_{i}(\Lambda_{i}, y_{i}) \) represents the concatenated feature vector derived from all LLM outputs for a given sample news article \( x_{i} \), and \( \theta \) denotes the parameter vector. The estimate of the true parameter denoted as \( \hat{\theta} \), is obtained by minimizing the negative marginal likelihood using the observed label matrix \( \Lambda \), relying solely on the patterns of agreement and disagreement within \( \Lambda \) since the true labels are unavailable. 

The estimated parameters are then used to produce probabilistic noisy labels: \( \hat{Y} = P_{\hat{\theta}}(Y \mid \Lambda) \), which can optionally serve as training labels for downstream models.

\item \textbf{\acrshort{wawa}}:
The \acrshort{wawa} algorithm \cite{ustalov2021learning} aggregates noisy labels through a three-step process. It begins by determining the \acrshort{majority_vote} label for each task. Then, it assesses each worker's skill by measuring the proportion of their responses that align with the \acrshort{majority_vote}. Finally, it computes a weighted \acrshort{majority_vote}, where the weights are derived from these skill estimates. The algorithm is based on the premise that responses agreeing with the \acrshort{majority_vote} are more likely to be accurate, and that a worker's reliability can be inferred from the consistency of their answers with the majority.

\item \textbf{Dawid-Skene}:
The Dawid-Skene aggregation model \cite{dawid1979maximum} is a probabilistic method used to estimate the expertise level of workers (in this case \acrshort{llm}s) in a crowd-sourcing environment. This model evaluates the reliability of each worker by constructing a confusion matrix that captures the probability of correctly or incorrectly labeling. The confusion matrix provides insight into how consistently a worker assigns labels in a classification problem.

In addition to individual worker reliability, Dawid-Skene incorporates a prior class probability vector, which represents the likelihood of each class occurring before considering the responses provided by workers. Since the true label of a task is unknown, it is treated as a latent variable that the model seeks to estimate. The probability of a task belonging to a specific class is influenced by the prior probabilities and the observed responses from workers. The optimization of these parameters, including the prior class probabilities, worker confusion matrices, and latent variables representing true labels, is achieved using the expectation maximization  algorithm.

\end{itemize}

\subsubsection{Discriminative Model}
The aggregated labels were used to fine-tune pre-trained \acrshort{llm}s: \acrshort{bert} and \acrshort{roberta}, and to train an \acrshort{lstm} model.
For pre-trained \acrshort{llm}s, this was achieved by adding a feedforward neural network and a classification head to the last layers while keeping the other pre-trained layers frozen to facilitate adaptation to our downstream \acrshort{epu} classification task. 

For an \acrshort{lstm}, the news article text was first tokenized and converted into sequences of integers. These sequences were then mapped into dense vector representations using an embedding layer, which captures semantic relationships between words. Initially, these embeddings are randomly initialized with a uniform distribution and are progressively updated during training. The transformed embeddings are then fed into the \acrshort{lstm} layers, which learns to identify sequential dependencies and contextual relationships within the text.

For both \acrshort{llm}s and an \acrshort{lstm}, the final layer consists of a dense network with a softmax function, which assigns the text to the appropriate \acrshort{epu} category. The training process optimizes the model by minimizing the expected loss over noisy labels formulated in Equation \ref{eqn:snorkel_gen_2} \cite{snorkel2017}. 
\begin{equation}\label{eqn:snorkel_gen_2}
    \hat{\theta} = \arg\min_{\theta} \sum_{i=1}^{n} \mathbb{E}_{\hat{y_{i}} \sim \hat{Y}} [\mathcal{L}(h_{\theta}(x_{i}), \hat{y_{i}})] 
\end{equation}
where \( \hat{\theta} \) represents the estimated model parameters, \( n \) is the number of training examples, \( \mathbb{E} \) denotes expectation, \( \mathcal{L} \) is the  expected loss with respect to $\hat{Y}$, \( h_{\theta} (x_{i}) \) is the output predicted by the discriminative model, \( x_{i} \) is a training instance, and \(\hat{y_{i}}  \) is the corresponding noisy probabilistic label from \( \hat{Y} \).

\subsubsection{Multi-label Classification for EPU Articles}  \label{sec:multi_label_method}
For articles positive for \acrshort{epu}, this section discusses the methodology used to identify specific subcategories, with their class distribution detailed in Table~\ref{tab:total_policy_labels_epu}. In contrast to the binary \acrshort{epu} classification task, where each article receives a single label, the subcategories here may simultaneously belong to multiple classes selected from a predefined set: \( \{a, b, \ldots, p\} \). Each article is a multi-hot vector \( s_{i} \), indicating the presence or absence of each corresponding subcategory.

In this section, experiments relied solely on human labeled data rather than synthetic labels. This decision stemmed from the absence of detailed guidelines in the labeling guide for assigning multiple labels to news articles, unlike the case with \acrshort{epu} classification. Consequently, it was challenging to develop clear and effective instructions for a reliable prompt-based method comparable to that used in \acrshort{epu} classification.

We used the same models as in \acrshort{epu} classification: \acrshort{bert}, \acrshort{roberta}, and \acrshort{lstm}. The text processing approach remained similar to that used for \acrshort{epu} classification, but the subcategories were converted into multi-hot vectors, denoted as $s_{i}$.  
Both models were trained using binary \acrshort{ce} loss across all subcategories defined in Equation \ref{eqn:multi_label_usa}.

\begin{equation}
\label{eqn:multi_label_usa}
    \mathcal{L}_{\text{multi}}(\mathbf{s}_i,\mathbf{\hat{s}}_i ) = - \frac{1}{n} \sum_{i=1}^{n} \sum_{j=1}^{m} \Bigl[s_{ij}\log(\hat{s}_{ij}) + (1 - s_{ij})\log\bigl(1 - \hat{s}_{ij}\bigr)\Bigr]
\end{equation}

where \( s_{ij} \) represents the true label for subcategory \( j \) of instance \( i \), where 1 indicates that the article belongs to subcategory \( j \) and 0 indicates it does not. The term \( \hat{s}_{ij} \) is the predicted probability that the article belongs to subcategory \( j \), outputted by the model.

\subsubsection{Hierarchical Classification for EPU Articles}\label{sec:hierarchical_method}  
Beyond multi-label classification, we performed hierarchical classification, where the classification task follows a structured dependency between an article’s general classification as \acrshort{epu}-related and its assignment to multiple subcategories. Unlike standard multi-label classification, where labels are treated as independent, hierarchical classification ensures that subcategories are only assigned when an article is identified as \acrshort{epu}-related. 

The dataset is now structured as: \(\{(x_i, y_i, \mathbf{s}_i)\}_{i=1}^{n} \), where \( x_i \) represents a news article, \( y_i \in \{0,1\} \) is a binary label indicating whether the article discusses \acrshort{epu}, and \( \mathbf{s}_i \) is a multi-hot vector indicating the presence or absence of multiple subcategories. 
We experimented with various hierarchical classification approaches: basic summation, masking, weighted, and a two-stage, discussed next.
 
\begin{itemize}\label{section:hierchical_approches}
    \item \textbf{Basic Summation}:  
A single model is trained to predict both \( y_i \) and \( \mathbf{s}_i \) simultaneously. It consists of a shared encoder followed by two classification heads: one for binary \acrshort{epu} classification and another for multi-label subcategory classification. For a given input text \( x_i \), the model processes it through the shared encoder and applies separate linear classifiers for each prediction. The loss function used for training is defined in Equation \ref{eqn:basic_summation}.
    \begin{equation}
    \label{eqn:basic_summation}
        \mathcal{L}_{\text{basic}}(y_i,\hat{y}_i, \mathbf{s}_i,\mathbf{\hat{s}}_i ) = \mathcal{L}_{\text{binary}}(y_i, \hat{y}_i) + \mathcal{L}_{\text{multi}}(\mathbf{s}_i, \mathbf{\hat{s}}_i),
    \end{equation}

    where \( \mathcal{L}_{\text{multi}} \) is a mult-label loss, and \( \mathcal{L}_{\text{binary}} \) is a binary \acrshort{ce} loss for predicting \( y_i \) defined in Equation \ref{eqn:binary_classification}.
    \begin{equation}
    \label{eqn:binary_classification}
    \mathcal{L}_{\text{binary}}(y_i,\hat{y}_i) = - \frac{1}{n} \sum_{i=1}^{n} \big[y_i \log \hat{y}_i + (1 - y_i) \log (1 - \hat{y}_i) \big]
    \end{equation}
    
    \item \textbf{Masking}:  
To enforce dependency constraints between binary and multi-label classification, this approach incorporates a masking mechanism. The loss function for subcategory prediction is applied only when \( y_i = 1 \), that is when the article is classified as \acrshort{epu}-related. The modified loss function is defined in Equation \ref{eqn:masking}.

    \begin{equation}
    \label{eqn:masking}
        \mathcal{L}_{\text{masking}}(y_i,\hat{y}_i, \mathbf{s}_i,\mathbf{\hat{s}}_i) = \mathcal{L}_{\text{binary}}(y_i, \hat{y}_i) + \mathbbm{1}[y_i = 1] \cdot \mathcal{L}_{\text{multi}}(\mathbf{s}_i, \mathbf{\hat{s}}_i),
    \end{equation}

    where \( \mathbbm{1}[y_i = 1] \) is an indicator function that applies the multi-label loss only to articles classified as \acrshort{epu}.

    \item \textbf{Weighted}:  
    Building upon the masking-based approach, this method further modifies the multi-label loss by introducing a weighting scheme to handle cases where articles are mistakenly classified as non-\acrshort{epu} but contain some subcategory information. Instead of completely ignoring subcategory loss for \( y_i = 0 \), different penalty weights are assigned for false positives and false negatives. The weighted loss function $\mathcal{L}_{\text{W}}$ is defined in Equation \ref{eqn:weighted_penalty}.
    \begin{equation}
    \label{eqn:weighted_penalty}
        \mathcal{L}_{\text{W}}(y_i,\hat{y}_i, \mathbf{s}_i,\mathbf{\hat{s}}_i) = \mathcal{L}_{\text{binary}}(y_i, \hat{y}_i) + \lambda_{\text{pos}} \mathbbm{1}[y_i = 1] \mathcal{L}_{\text{multi}}(\mathbf{s}_i, \mathbf{\hat{s}}_i) + \lambda_{\text{neg}} \mathbbm{1}[y_i = 0] \mathcal{L}_{\text{multi}}(\mathbf{s}_i, \mathbf{\hat{s}}_i)
    \end{equation}

    where \( \lambda_{\text{pos}} \) and \( \lambda_{\text{neg}} \) are hyperparameters controlling the relative penalty for subcategory errors when \( y_i \) is 1 versus 0.

    \item \textbf{Two-stage}:  
Instead of jointly predicting \( y_i \) and \( \mathbf{s}_i \), this approach separates the task into two stages: binary classification and multi-stage classification, trained separately unlike in the masking approach. 

\textit{Binary Classification Stage}:  
The first stage is a binary classifier trained solely on predicting whether an article discusses \acrshort{epu}. The model is optimized using the \acrshort{ce} loss \(\mathcal{L}_{\text{binary}}(y_i,\hat{y}_i)\) defined in Equation \ref{eqn:binary_classification}.

\textit{Multi-Label Classification Stage}:  
After training the binary classifier, its predicted labels \( \hat{y}_i \) are used to select articles for the second-stage multi-label classifier, which is applied only to samples where \( \hat{y_i} = 1 \). The loss function \(\mathcal{L}_{\text{multi}}( \mathbf{s}_i,\mathbf{\hat{s}}_i )\) is the same as in standard multi-label classification but limited to this filtered subset.

\end{itemize}

\subsection{Data}
We used a dataset of $12,000$ news articles provided by Baker et al. (2016), selected from \acrshort{usa} news articles that included the terms “economy” and “uncertainty”. Domain experts assigned binary labels to these articles based on whether they discussed \acrshort{epu}.  
Among these, \( 4939 \) were identified as related to \acrshort{epu}. These articles were then classified into one or more multi-label subcategories, which explains why the total count in the class distribution shown in Table \ref{tab:total_policy_labels_epu} is \( 6037 \) instead of \( 4939 \), as some articles fall into multiple subcategories.

The subcategories include: government spending (entitlement programs, social safety nets, welfare programs), trade policy (tariffs, quotas, trade rules), financial regulation (banking and equity markets), energy and environmental regulation (natural resources, commodities), sovereign debt (exchange rate policy, foreign reserves), political conflict and leadership changes (elections, governance crises), and legal policy (judicial decisions, regulatory changes).  
Further details about the dataset can be obtained from the authors' website\footnote{\url{https://www.policyuncertainty.com}} \cite{baker2016measuring}.

\begin{table}[h]
\centering
\caption{Distribution of subcategories for news articles describing \acrshort{epu}}
\label{tab:total_policy_labels_epu}
\renewcommand{\arraystretch}{1.2} 
\setlength{\tabcolsep}{1em} 
\begin{tabularx}{\textwidth}{|X|r|}
\toprule
\textbf{Subcategory} & \textbf{Count} \\
\midrule
Other policy matters that do not fit into pre-defined categories & 1176 \\
Political conflict and leadership changes & 827 \\
National security and terrorism & 820 \\
Trade policy (tariffs, quota, trade rules) & 712 \\
Taxes & 659 \\
Monetary policy & 372 \\
Government spending & 367 \\
Energy \& environmental regulation, natural resources and commodities & 187 \\
Competition policy & 182 \\
Labor regulations & 166 \\
Legal policy & 164 \\
Sovereign debt, exchange rate policy, foreign reserves & 127 \\
Fiscal policy & 125 \\
Financial regulation (including banking and equity markets) & 100 \\
Entitlement programs, social safety net, welfare programs & 46 \\
Health care programs and regulations & 7 \\
\midrule
\textbf{Grand Total} & \textbf{6037} \\
\bottomrule
\end{tabularx}
\end{table}

\subsection{Experimental Setup and Evaluation} 
\label{sec:setup_weak_supervision}
All experiments were conducted in Python with \href{google-bert/bert-base-uncased}{\acrshort{bert}} and \href{FacebookAI/roberta-base}{\acrshort{roberta}} being implemented using the Huggingface  Transformer library~\footnote{\url{https://huggingface.co/}}. \acrshort{lstm} \cite{hochreiter1997long} was implemented using the Keras library~\footnote{\url{https://keras.io/}} \cite{chollet2018keras}. 

For the hyperparameters, the maximum text length was set to \(512\) tokens, the Adam optimizer was used, with a learning rate of \(1 \times 10^{-3}\), a batch size of \(50\), and training was performed over \(10\) epochs.  
All model experiments were conducted on a single A100 40GB \acrshort{gpu} machine.

For synthetic label generation, we primarily utilized generative \acrshort{llm}s, specifically open-source models from the \acrshort{llama}, Mistral, and Qwen families, using the prompt shown in Figure \ref{fig:data_generation_prompt_weak}. The selected models represented the \acrshort{sota} at the time of the experiments, that we could perfom inference on with the hardware that was available to us.
The specific versions used in the experiments include: \href{mistralai/Mistral-Small-24B-Instruct-2501}{\acrshort{mistral_24b}}, \href{Qwen/Qwen2.5-7B-Instruct-1M}{\acrshort{qwen2_7b}}, \href{meta-llama/Llama-3.1-8B-Instruct}{\acrshort{llama3_8b}}, \href{mistralai/Mistral-7B-Instruct-v0.3}{\acrshort{mistral3_7b}}, and \href{Qwen/Qwen2.5-14B-Instruct}{\acrshort{qwen2_14b}}.
 
Binary classification performance was evaluated using accuracy, F1 , \acrlong{gm} (\acrshort{gm}), \acrlong{auc} (\acrshort{auc}), and \acrlong{auprc} (\acrshort{auprc}). Accuracy represents the overall proportion of correctly classified instances. F1 is the harmonic mean of precision and recall. \acrshort{gm} accounts for both sensitivity and specificity, suitable under class imbalance setting. \acrshort{auc} measures the model’s ability to distinguish between classes across different threshold settings, while \acrshort{auprc} assesses precision-recall trade-offs, making it particularly relevant for imbalanced data.

Multi-label classification performance was further assessed using \acrlong{emr} (\acrshort{emr}) and \acrlong{lrp} (\acrshort{lrp}) \cite{zhang2013review}. \acrshort{emr} calculates the percentage of instances where the predicted set of labels exactly matches the true set. \acrshort{lrp} evaluates how well the model ranks correct labels higher than incorrect ones, focusing on the order of label relevance rather than requiring an exact match. 

In addition to classification metrics, we used \acrlong{ck} (\acrshort{ck}) and \acrlong{mcc} (\acrshort{mcc}) to measure the correlation and agreement of different label generators with humans. \acrshort{ck} \cite{hsu2003interrater} evaluates the level of agreement between the predicted and actual labels while accounting for the possibility of random agreement.
\acrshort{mcc} \cite{chicco2023matthews}, on the other hand, provides a balanced measure of correlation between predictions and ground truth, even when class distributions are uneven by considering all elements of the confusion matrix.

The code, datasets, and hyperparameters referenced in this research are available in the GitHub repository: \url{https://github.com/TrustPaul/Weak-supervision-in-Economic-Policy-Uncertanity}.

\section{Results and Discussion}\label{sec: results}
This section presents the results of the various experiments for binary \acrshort{epu} classification, as well as multi-label and hierarchical classification.

\begin{table}[h]
\centering
\caption{Evaluation metrics (\acrshort{ck} and \acrshort{mcc}) for different label generators for the binary \acrshort{epu} classification task}
\label{tab:evaluation_metrics_mcc_ck}
\renewcommand{\arraystretch}{1.2} 
\setlength{\tabcolsep}{1.5em} 
\begin{tabularx}{\textwidth}{|X|c|c|}
\toprule
\textbf{Label Generator} & \textbf{\acrshort{ck}} & \textbf{\acrshort{mcc}} \\
\midrule
\acrshort{majority_vote} & -0.0121 & -0.0139 \\
\acrshort{wawa} & -0.0011 & -0.0013 \\
\acrshort{mistral3_7b} & 0.0460 & 0.0553 \\
\acrshort{llama3_8b} & 0.2970 & 0.3058 \\
\acrshort{qwen2_14b} & 0.2849 & 0.3486 \\
\acrshort{qwen2_7b} & 0.2849 & 0.3486 \\
\acrshort{mistral_24b} & 0.3537 & 0.3766 \\
\acrshort{dp} & 0.4156 & 0.4156 \\
\textbf{Dawid-Skene} & \textbf{0.4813} & \textbf{0.4908} \\
\bottomrule
\end{tabularx}
\end{table}

\textbf{Analysis of Different Label Generators}: Table \ref{tab:evaluation_metrics_mcc_ck} presents evaluation metrics (\acrshort{mcc} and \acrshort{ck}) comparing the performance of models of different label generators. 
The evaluation includes individual \acrshort{llm}s such as \acrshort{mistral_24b}, \acrshort{qwen2_14b}, \acrshort{qwen2_7b}, \acrshort{llama3_8b}, and \acrshort{mistral3_7b}, as well as aggregation methods like Dawid-Skene, \acrshort{wawa}, and \acrshort{majority_vote}. Dawid-Skene achieved the highest correlation with human labels, with an \acrshort{mcc} of 0.4908 and a \acrshort{ck} of 0.4813, followed by the \acrshort{dp} method, which obtained an \acrshort{mcc} and \acrshort{ck} of 0.4156. 

Among individual \acrshort{llm}s, \acrshort{mistral_24b} showed the strongest correlation with human labels (\acrshort{mcc} = 0.3766, \acrshort{ck} = 0.3537), while \acrshort{majority_vote} had the lowest correlation. These results indicate that proper aggregation or the use of more effective models leads to higher agreement with human labels, and the high values of \acrshort{mcc} and \acrshort{ck} observed especially by top methods like Dawid-Skene, \acrshort{dp} and \acrshort{mistral_24b} suggest that these labels can be reliably used for training other \acrshort{llm}s for this task.

\begin{table}[h]
\centering
\caption{Performance evaluation metrics (F1, accuracy, and \acrshort{gm}) for different label generators for the binary \acrshort{epu} classification task}
\label{tab:evaluation_metrics_untrained_llm}
\renewcommand{\arraystretch}{1.2} 
\setlength{\tabcolsep}{1.5em} 
\begin{tabularx}{\textwidth}{|X|c|c|c|}
\toprule
\textbf{Label Generator} & \textbf{Accuracy} & \textbf{F1} & \textbf{\acrshort{gm}} \\
\midrule
\acrshort{mistral3_7b}   & 42.06 & 27.36 & 41.29 \\
\acrshort{majority_vote} & 55.25 & 46.65 & 49.44 \\
\acrshort{wawa}          & 55.41 & 47.69 & 49.95 \\
\acrshort{llama3_8b}     & 64.15 & 42.85 & 56.09 \\
\acrshort{qwen2_14b}     & 68.89 & 61.46 & 62.85 \\
\acrshort{qwen2_7b}      & 68.89 & 61.46 & 62.85 \\
Dawid-Skene              & 75.01 & 49.30 & 63.12 \\
\acrshort{mistral_24b}   & 70.72 & 66.80 & 66.61 \\
\textbf{\acrshort{dp}}   & \textbf{71.83} & \textbf{70.78} & \textbf{70.77} \\
\bottomrule
\end{tabularx}
\end{table}

Beyond correlation and agreement metrics, classification performance is of greater interest for this task. Table \ref{tab:evaluation_metrics_untrained_llm} therefore presents the performance of various data generators without additional training. For \acrshort{llm}s, this corresponds to zero-shot prompting performance. \acrshort{dp} achieved the highest performance results, with a \acrshort{gm} of \(70.77\%\) and an F1 score of \(70.78\%\), followed by \acrshort{mistral_24b}, which obtained a \acrshort{gm} of \(66.61\%\) and an F1 score of \(66.80\%\). The lowest performance was observed with \acrshort{mistral3_7b} and the \acrshort{majority_vote} method. Based on these results, labels generated by \acrshort{dp} were selected as synthetic labels for training other models.

The findings indicate that ensembling improves performance, but the choice of aggregation method significantly impacts the outcome. The zero-shot performance of \acrshort{llm}s like \acrshort{mistral_24b} suggests that sufficiently large and high-performing models can achieve reasonable classification results through prompting alone. However, even in cases where zero-shot prompting performs well, achieving such results requires models with at least 7 billion parameters, which come with significant inference costs. Therefore, it remains beneficial to use labels generated by these \acrshort{llm}s or aggregation methods in a distillation process to train smaller models, such as \acrshort{bert}, which offer lower training and inference costs.

\textbf{Fine-tuning results for binary classification of \acrshort{epu}}:
Table \ref{tab:model_performance_epu_usa} presents the performance of different models along with their training data sources, which include human-labeled data and \acrlong{ws} (\acrshort{ws}) where labels are generated through aggregation. The models compared are \acrshort{lstm}, \acrshort{bert}, and \acrshort{roberta}.  

In general, models perform better when trained on human labeled data. This is particularly evident with \acrshort{bert}, which achieves an F1 score of \(74.66\%\) and a \acrshort{gm} of \(73.77\%\) when trained on human labeled data, compared to an F1 score of \(71.92\%\) and a \acrshort{gm} of \(71.66\%\) when trained on synthetic labels. A similar trend is observed with \acrshort{roberta}, which performs almost as well as \acrshort{bert} but achieves slightly higher F1 and \acrshort{gm} scores when trained on synthetic data. \acrshort{lstm} has the lowest performance on both datasets, showing only minor differences in F1 and \acrshort{gm} scores between human labeled and synthetic data.

\begin{table}[h]
\centering
\caption{Performance comparison of different models and datasets on \acrshort{epu} binary classification. Dataset \acrshort{ws} indicates that labels generated by \acrshort{dp} were used to train the respective model.}
\label{tab:model_performance_epu_usa}
\renewcommand{\arraystretch}{1.2} 
\setlength{\tabcolsep}{1em} 
\begin{tabularx}{\textwidth}{|c|X|c|c|c|c|c|}
\toprule
\textbf{Dataset} & \textbf{Model} & \textbf{Accuracy} & \textbf{F1} & \textbf{\acrshort{gm}} & \textbf{\acrshort{auc}} & \textbf{\acrshort{auprc}} \\
\midrule
Human & \acrshort{bert}    & \textbf{75.99} & \textbf{74.66} & \textbf{73.77} & 0.8423 & 0.7750 \\
Human & \acrshort{roberta} & 74.97 & 73.54 & 72.59 & \textbf{0.8440} & \textbf{0.7833} \\
Human & \acrshort{lstm}    & 63.41 & 60.79 & 58.96 & 0.6452 & 0.5352 \\
\acrshort{ws} & \acrshort{dp}      & 71.83 & 70.78 & 70.55 & 0.7338 & 0.6654 \\
\acrshort{ws} & \acrshort{lstm}    & 63.46 & 62.10 & 61.66 & 0.6482 & 0.5359 \\
\acrshort{ws} & \acrshort{bert}    & 71.93 & 71.33 & 71.83 & 0.7859 & 0.6877 \\
\acrshort{ws} & \acrshort{roberta} & 73.03 & 71.92 & 71.66 & 0.7936 & 0.7114 \\
\bottomrule
\end{tabularx}
\end{table}

The results in Table \ref{tab:model_performance_epu_usa} further reveal that models trained on synthetically generated data such as \acrshort{roberta}, outperform the zero-shot prompting performance of the \acrshort{llm} used to generate the synthetic data. \acrshort{roberta} achieves a higher \acrshort{gm} and F1 score than \acrshort{dp} and also exceeds the best zero-shot prompting \acrshort{llm}: \acrshort{mistral_24b}, in terms of \acrshort{gm} (\(71.66\%\) versus \(66.61\%\)). 

Although the performance of zero-shot prompting and aggregation methods is close to that of fine-tuned models, the competitive results are mainly achieved using extremely large \acrshort{llm}s, as previously noted.  
Despite this, using \acrshort{llm}s to generate synthetic labels for training smaller models remains a promising approach, as human annotation is costly and time-consuming, making synthetic data generation a more efficient and scalable alternative.

  \begin{figure}[H]
	\centering
		\includegraphics[width=140mm]{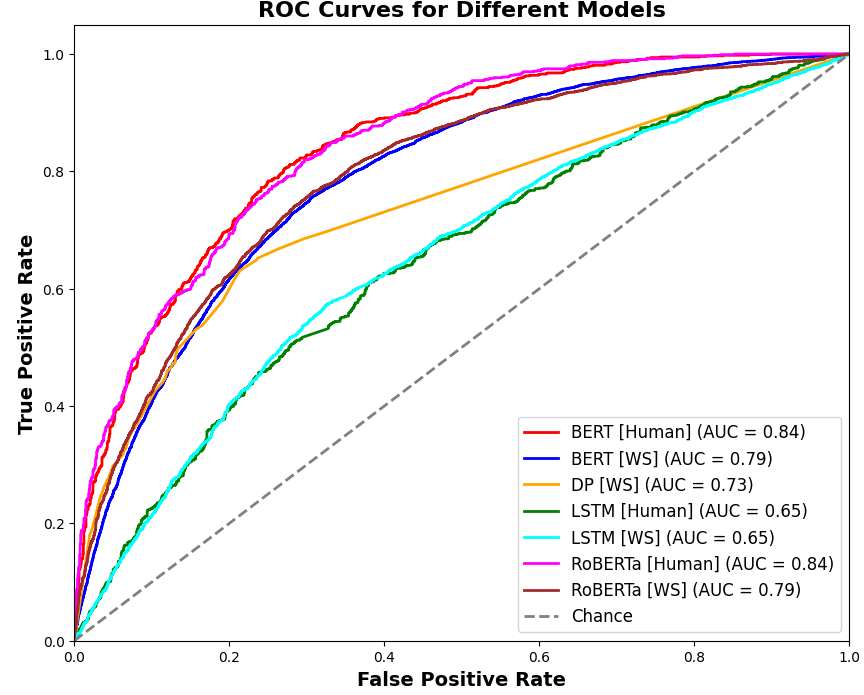}
     \caption{\acrlong{roc} (\acrshort{roc}) curves for  different models for binary \acrshort{epu} classification task. \acrshort{ws} in the case means that labels generated by \acrshort{dp} were used to train the respective model. }
     \label{fig:roc_usa_unweighted}
\end{figure}
In addition to the classification metrics in Table \ref{tab:model_performance_epu_usa}, Figure \ref{fig:roc_usa_unweighted} presents the \acrshort{roc} curves for different models.
The trends in the \acrshort{roc} align with those observed in the F1 and \acrshort{gm} scores. Models trained on human labeled data generally achieve higher \acrshort{auc} values, with both \acrshort{roberta} and \acrshort{bert} reaching approximately \(0.84\). In contrast, \acrshort{lstm} shows consistently lower \acrshort{auc} values, with little difference between models trained on weak supervision and those trained on human labeled data.

\begin{table}[h]
\centering
\caption{Performance metrics for different models on the multi-label text classification task}
\label{tab:model_performance_multi_label_epu}
\renewcommand{\arraystretch}{1.2} 
\setlength{\tabcolsep}{1.5em} 
\begin{tabularx}{\textwidth}{|X|c|c|c|c|}
\toprule
\textbf{Model} & \textbf{\acrshort{emr}} & \textbf{\acrshort{lrp}} & \textbf{\acrshort{auprc}} & \textbf{\acrshort{auc}} \\
\midrule
\acrshort{lstm}    & 58.15 & 75.40 & 0.0466 & 0.5931 \\
\acrshort{bert}    & \textbf{61.70} & 83.70 & 0.1634 & 0.7727 \\
\acrshort{roberta} & 60.43 & \textbf{83.78} & \textbf{0.1676} & \textbf{0.7930} \\
\bottomrule
\end{tabularx}
\end{table}

\textbf{Multi-Label Text Classification}: For \acrshort{epu} binary classification, articles identified as positive for \acrshort{epu} were further assigned to 16 multi-label categories, with the class distribution detailed in Table \ref{tab:total_policy_labels_epu}. For multi-label classification, only human labeled data was used for fine-tuning. The models evaluated include: \acrshort{roberta}, \acrshort{bert}, and \acrshort{lstm}, assessed using multi-label evaluation metrics such as \acrshort{emr}, \acrshort{lrp}, \acrshort{auc}, and \acrshort{auprc} shown in Table \ref{tab:model_performance_multi_label_epu}. 

In terms of \acrshort{emr}, all three models exhibited comparable performance, with \acrshort{bert} achieving the highest score. \acrshort{lstm} recorded an \acrshort{emr} of $58.15\%$, while \acrshort{bert} and \acrshort{roberta} attained $61.70\%$ and $60.43\%$, respectively. For the \acrshort{lrp} metric, \acrshort{bert} and \acrshort{roberta} performed similarly, both reaching approximately $83.70\%$, whereas \acrshort{lstm} had a lower \acrshort{lrp} score of $75.40\%$.

The \acrshort{auprc} values were generally low across all models, with \acrshort{roberta} obtaining the highest \acrshort{auprc} and \acrshort{lstm} the lowest. In terms of \acrshort{auc}, \acrshort{roberta} achieved the highest value of $0.7930$, while \acrshort{lstm} had the lowest at $0.5931$. These results indicate that pre-trained \acrshort{llm}s such as \acrshort{bert} and \acrshort{roberta} perform better in multi-label classification compared to \acrshort{lstm}. However, their performance is significantly lower than in binary classification, which is expected since multi-label classification is a more complex task.

\begin{table}[h]
\centering
\caption{Performance evaluation of various models and hierarchical strategies on the binary \acrshort{epu} classification task}
\label{tab:performance_hierchical_EPU}
\renewcommand{\arraystretch}{1.2} 
\setlength{\tabcolsep}{1em} 
\begin{tabularx}{\textwidth}{|l|X|c|c|c|c|c|}
\toprule
\textbf{Model} & \textbf{Hierarchical Strategy} & \textbf{Accuracy} & \textbf{F1} & \textbf{\acrshort{auc}} & \textbf{\acrshort{auprc}} & \textbf{\acrshort{gm}} \\
\midrule
\multirow{4}{*}{\acrshort{bert}} 
 & basic     & 74.81 & 74.33 & 83.86 & 0.7727 & 74.54 \\
 & masking   & 74.72 & 74.30 & 84.16 & 0.7724 & 74.58 \\
 & two-stage & 74.64 & 74.33 & 83.80 & 0.7641 & 74.80 \\
 & weighted  & 75.62 & 74.88 & 83.93 & 0.7741 & 74.90 \\
\midrule
\multirow{4}{*}{\acrshort{roberta}} 
 & basic     & 76.03 & 75.48 & 84.68 & 0.7895 & 75.60 \\
 & masking   & \textbf{76.19} & \textbf{75.59} & \textbf{84.81} & \textbf{0.7905} & 75.66 \\
 & two-stage & 76.15 & 75.40 & \textbf{84.81} & 0.7891 & 75.41 \\
 & weighted  & 75.91 & 75.22 & 84.15 & 0.7801 & 75.25 \\
\midrule
\multirow{4}{*}{\acrshort{lstm}} 
 & basic     & 61.62 & 58.66 & 61.93 & 0.5069 & 59.08 \\
 & masking   & 62.31 & 57.93 & 61.03 & 0.5023 & 59.21 \\
 & two-stage & 61.86 & 57.95 & 59.18 & 0.4644 & 58.89 \\
 & weighted  & 62.15 & 56.15 & 60.56 & 0.4967 & 58.88 \\
\bottomrule
\end{tabularx}
\end{table}

\textbf{Hierarchical \acrshort{epu} Classification}: Since the task was hierarchical, where articles classified as \acrshort{epu} could be further assigned to 16 multi-label categories, we investigated multiple strategies for hierarchical text classification. Details on these methods are provided in Section \ref{sec:hierarchical_method} of the methodology.  

The strategies include a basic approach, where the model predicts both binary and multi-label classifications simultaneously, a masking approach, where multi-label predictions are made only for articles classified as \acrshort{epu}, a weighted approach, which applies an additional loss penalty for misclassifying articles as not containing \acrshort{epu}, and a two-stage approach, where two models are trained separately, one for binary classification and another for multi-label classification. 

Table \ref{tab:performance_hierchical_EPU} presents the binary \acrshort{epu} classification results for different models across these approaches. Overall, performance differences among the four methods are minimal across all models and evaluation metrics. Among the models, \acrshort{roberta} achieved the highest \acrshort{gm} of $75.66\%$ using the masking approach. \acrshort{bert} performed best with the two-stage approach, reaching a \acrshort{gm} of $74.80\%$. For \acrshort{lstm}, the highest \acrshort{gm} was obtained using the masking approach.

\begin{table}[h]
\centering
\caption{Performance evaluation of various models and hierarchical strategies on the multi-label classification task}
\label{tab:performance_hierchical_epu_multilabel}
\renewcommand{\arraystretch}{1.2} 
\setlength{\tabcolsep}{1em} 
\begin{tabularx}{\textwidth}{|l|X|c|c|c|c|}
\toprule
\textbf{Model} & \textbf{Hierarchical Strategy} & \textbf{\acrshort{emr}} & \textbf{\acrshort{auprc}} & \textbf{\acrshort{auc}} & \textbf{\acrshort{lrp}} \\
\midrule
\multirow{4}{*}{\acrshort{bert}} 
 & basic     & 59.45 & 0.0702 & 0.7080 & 74.41 \\
 & masking   & 59.45 & 0.0811 & 0.7189 & 74.10 \\
 & two-stage & \textbf{61.94} & 0.1974 & 0.7526 & \textbf{81.21} \\
 & weighted  & 59.45 & 0.0678 & 0.7046 & 73.53 \\
\midrule
\multirow{4}{*}{\acrshort{roberta}} 
 & basic     & 59.45 & 0.0765 & 0.7070 & 74.39 \\
 & masking   & 59.45 & 0.0963 & 0.7219 & 75.22 \\
 & two-stage & 61.66 & \textbf{0.2231} & \textbf{0.7745} & 81.02 \\
 & weighted  & 59.45 & 0.0679 & 0.6999 & 73.48 \\
\midrule
\multirow{4}{*}{\acrshort{lstm}} 
 & basic     & 59.45 & 0.0395 & 0.5879 & 68.48 \\
 & masking   & 59.45 & 0.0399 & 0.5807 & 67.55 \\
 & two-stage & 59.25 & 0.0599 & 0.5919 & 67.77 \\
 & weighted  & 59.45 & 0.0377 & 0.5620 & 66.90 \\
\bottomrule
\end{tabularx}
\end{table}

\textbf{Hierarchical Multi-label Text Classification}: Table \ref{tab:performance_hierchical_epu_multilabel} presents the \acrshort{emr}, \acrshort{lrp}, \acrshort{auprc}, and \acrshort{auc} metrics for the three models across the four hierarchical approaches. The \acrshort{emr} values remain nearly identical across all approaches.  
For \acrshort{lrp}, the best performance is achieved by \acrshort{bert} using the two-stage approach, followed by \acrshort{roberta} again under the same approach with an \acrshort{lrp} of $81.02\%$. \acrshort{lstm} has consistently lower \acrshort{lrp} values overall, but its best result is obtained with the basic approach at $68.48\%$. \acrshort{roberta} achieves the highest \acrshort{auc} value of $0.7745$ using the two-stage approach.

\section{Conclusion} \label{section: conclusion}
This research investigated the application of \acrshort{llm}s for \acrshort{epu} classification, addressing limitations of earlier approaches based on keyword lists, unsupervised \acrshort{ml}, or supervised methods dependent on costly human annotations. To reduce reliance on manual labeling, we employed weak supervision by generating labels with \acrshort{llm}s, where large models such as \acrshort{mistral_24b} achieved the strongest performance in zero-shot prompting for label generation, and aggregation methods like \acrshort{dp} and Dawid–Skene further enhanced label quality. While models trained on human labeled data achieved the highest accuracy, the gap to weakly supervised models was relatively small, indicating that \acrshort{llm}-based labeling offers a practical and scalable alternative. Finally, extending beyond binary detection, we examined multi-label and hierarchical classification, finding that fine-tuned pre-trained \acrshort{llm}s delivered the best performance, with the two-stage strategy most effective for policy subcategory classification, though overall differences across hierarchical methods were minor.
... 

\bibliographystyle{plainnat}llll
\bibliography{sn-bibliography}

\end{document}